# Evaluation of direct attacks to fingerprint verification systems

**J. Galbally · J. Fierrez · F. Alonso-Fernandez ·
M. Martinez-Diaz**

**Abstract** The vulnerabilities of fingerprint-based recognition systems to direct attacks with and without the cooperation of the user are studied. Two different systems, one minutiae-based and one ridge feature-based, are evaluated on a database of real and fake fingerprints. Based on the fingerprint images quality and on the results achieved on different operational scenarios, we obtain a number of statistically significant observations regarding the robustness of the systems.

## 1 Introduction

Biometric systems present several advantages over classical security methods based on something that you know (PIN, Password, etc.) or something that you have (key, card, etc.) [1, 2]. Traditional authentication systems cannot discriminate between impostors who have illegally acquired the privileges to access a system and the genuine user. Furthermore,

J. Galbally (✉) · J. Fierrez · F. Alonso-Fernandez ·
M. Martinez-Diaz
Universidad Autonoma de Madrid, EPS, C/Francisco
Tomas y Valiente 11, 28049 Madrid, Spain
e-mail: javier.galbally@uam.es

J. Fierrez
e-mail: julian.fierrez@uam.es

F. Alonso-Fernandez
e-mail: fernando.alonso@uam.es

M. Martinez-Diaz
e-mail: marcos.martinez@uam.es

in biometric systems there is no need for the user to remember difficult PIN codes that could be easily forgotten or to carry a key that could be lost or stolen.

However, in spite of these advantages, biometric systems have some drawbacks [3], including: i) the lack of secrecy (e.g. everybody knows our face or could get our fingerprints), and ii) the fact that a biometric trait cannot be replaced (if we forget a password we can easily generate a new one, but no new fingerprint can be generated if an impostor "steals" it). Furthermore, biometric systems are vulnerable to external attacks which could decrease their level of security.

In [4] Ratha identified and classified eight possible attack points to biometric recognition systems. These vulnerability points, depicted in Fig. 1, can broadly be divided into two main groups:

– **Direct attacks**. In [4] the possibility to generate synthetic biometric samples (e.g., speech, fingerprints or face images) in order to illegally access a biometric system was discussed, corresponding to the attack point 1 depicted in Fig. 1. These attacks at the sensor level are referred to as *direct attacks*. It is worth noting that in this type of attacks no specific knowledge about the system is needed (e.g., matching algorithm used, feature extraction, feature vector format, etc.). Furthermore, the attack is carried out in the analog domain, outside the digital limits of the system, so the digital protection mechanisms (e.g., digital signature, watermarking, etc.) cannot be used.
– **Indirect attacks**. This group includes all the remaining seven points of attack identified in Fig. 1. Attacks 3 and 5 might be carried out using a Trojan Horse that bypasses the feature extractor, and the matcher respectively. In attack 6 the system database is manipulated (e.g., a template is modified, added or deleted) in order to gain access to the application. The remaining points of attack (2,

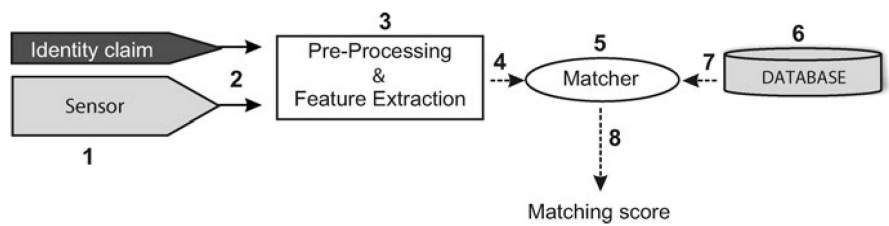

4, 7 and 8) are thought to exploit possible weak points in the communication channels of the system, extracting, adding or changing information from them. In opposition to direct attacks, in this case the intruder needs to have some additional information about the internal working of the recognition system and, in most cases, physical access to some of the application components (feature extractor, matcher, database, etc.) is required.

From a more general point of view, Maltoni et al. have classified in [5] the typical threats that may affect any security application, not only based on biometric recognition. Among all the possible attacks several are emphasized, namely: i) *Denial of Service* (DoS) where the attacker damages the system so that it can no longer be accessed by the legitimate users, ii) *circumvention*, in this case an unauthorized user gains access to the system, iii) *repudiation*, in this type of threat it is the legitimate user who denies having accessed the system, iv) *contamination or covert acquisition*, this is the case of the direct attacks presented before [4], v) *collusion*, in this attack a user with special privileges (e.g. administrator) allows the attacker to bypass the recognition component, vi) *coercion*, legitimate users are forced to help the attacker enter the system. For security systems based on biometric recognition the contamination and circumvention attacks can be identified respectively with the direct and indirect attacks previously presented.

In the present work we will focus in the system model shown in Fig. 1 to carry out a systematic and replicable evaluation of the vulnerabilities of fingerprint-based recognition systems to direct attacks. For this evaluation two general attack scenarios have been considered, namely: i) with a cooperative user, and ii) without the cooperation of the user. A database of real and fake fingerprints was specifically created for each of these two scenarios,[1] using three different sensors each belonging to one of the main technologies existing in the market: two flat (optical and capacitive), and one sweep sensor. Two different systems (the NIST minutiae-based and one proprietary ridge-based) are tested on these two databases and we present the results considering the normal operation mode (i.e., enrollment and test are carried out with real fingerprints) as reference and two different attacks, namely: i) enrollment and test are performed with fake fingerprints (attack 1), and ii) enrollment is carried out with real fingerprints while the test is done with the corresponding fake imitations (attack 2).

Previous related works have already studied direct attacks to fingerprint verification systems [8, 9], but usually with very small datasets, thus resulting in statistically insignificant results. The contributions of the present paper over previous works, which are mainly our vulnerability assessment methodology and various experimental findings, are based on a large set of data from diverse subjects and acquisition conditions. In particular, we show a strong correlation between the image quality of fake fingerprints and the robustness against direct attacks for minutiae-based verification systems, while the robustness against direct attacks of ridge-based systems is less affected by image quality. The results reported are therefore relevant to devise proper countermeasures against the considered attacks depending on the system at hand.

The paper is structured as follows. Related works are summarized in Sect. 2. The systems evaluated and the gummy fingers generation process are described in Sect. 3 and Sect. 4 respectively. The databases and experimental protocols followed are presented in Sect. 5. In Sect. 6 we describe and analyze the results obtained. Conclusions are finally drawn in Sect. 7.

## 2 Related works

It has been shown in several works, usually in very small datasets and not always in a systematic and replicable way, that a biometric system can be fooled by means of presenting a synthetic trait to the sensor.

The first effort in fingerprint spoofing can be traced back to the 1920s and was executed by Wehde [6], that used his knowledge in photography and engraving to generate gummy fingers from latent prints. Using forensic techniques the latent fingerprint was highlighted and a photograph taken. That picture was later used to engrave a copper plate that could be used to leave false latent fingerprints on objects.

---

[1] The real and fake fingerprint databases used in this paper will be available at http://atvs.ii.uam.es.

In modern times, one of the first published evaluations of fingerprint based recognition systems against spoofing methodologies was described in [7]. More recently, in [8] Putte and Keuning tested the vulnerability of several sensors to fake fingerprints made with plasticine and silicone. The authors classified the different methods to create gummy fingers in two main categories: with and without cooperation of the legitimate user. Then one method of each class was described and results on six commercial sensors (optical and solid state) were reported. Out of the six sensors tested five of them accepted the imitation as real on the first attempt while the remaining sensor permitted the access to the system on the second attempt.

Matsumoto et al. [9] carried out similar experiments to those reported in [8], this time with fake fingerprints made of gelatin. Again they distinguished between the case in which they had the cooperation of the fingerprint owner and the situation in which the latent fingerprint had to be lifted from a surface (i.e., latent fingerprints). When the genuine user cooperated to make the gummy fingers, recognition rates ranging from 68 to 100% were reported for the fake fingerprints in all the 11 systems tested. Regarding the imitations generated with non cooperative users, the acceptance rate was always above 60%. Similar experiments testing different sensors and using several attacking methods can be found in [10] and [11].

Recently, it has been shown in [12] that it is possible to carry out direct attacks using fingertips reconstructed from standard minutiae templates. In that work, the gummy fingers were not generated directly from the user's trait (cooperative method) or recovered from a latent fingerprint (noncooperative technique). In [12] the fingerprint image was first reconstructed from the user's minutiae template, and then the reconstructed image was used to generate the synthetic trait. Although the technique comprises two different stages (from the template to the image [13], and from the image to the gummy finger), and therefore some errors in the reconstruction process were accumulated, the attack showed an efficiency of over 70% for almost all the scenarios considered.

Recent works related to the vulnerabilities of fingerprint verification systems report liveness detection methods. These algorithms are anti-spoofing techniques which use different physiological properties to distinguish between real and gummy fingerprints, thus improving the robustness of the system against direct attacks. Liveness detection approaches can broadly be divided into: i) software-based techniques which try to detect fake fingerprints once the sample has been acquired with a standard sensor [14–17], and ii) hardware-based techniques in which some specific device is added to the sensor in order to detect particular properties of a living finger (e.g., blood pressure, odor, etc.) [18–21].

## 3 Fingerprint verification systems

Two different fingerprint verification systems, one minutiae-based and one ridge-based, are used in the experiments:

– The minutiae-based NIST Fingerprint Image Software 2 (NFIS2) [22]. It is a PC-based fingerprint processing and recognition system formed of independent software modules. The feature extractor generates a text file containing the location, orientation and quality of each minutia from the fingerprint. The matcher uses this file to generate the score. The matching algorithm is rotation and translation invariant since it computes only relative distances and orientations.

– A basic ridge-based fingerprint verification system [23]. Most of the actual fingerprint verification systems are minutiae-based as this is the basis of the fingerprint comparison made by fingerprint examiners. However, although minutiae may carry most of the fingerprint discriminant information, under certain circumstances (e.g. bad quality images) the extraction of a reliable minutiae map could be quite challenging. In this cases the use of some complementary recognition system, for example based on the ridge pattern, can improve the global performance of the system [24]. The system tested in this work uses 8 Gabor filters (each rotated $27.5°$ with respect to the previous one) to capture the ridge pattern. The 8 resulting images are tesselated in a rectangular grid and the variance of the filter responses in each cell are used to generate the feature vector. No rotation alignment is applied to the input images so it is quite sensitive to fingerprint rotation. For more details we refer the reader to [23].

## 4 Generation process of the gummy fingers

For each of the attack scenarios considered, with and without cooperation of the user, a database of real and fake fingers was created for the experiments. The generation process of the gummy imitations differs for each of the two scenarios:

– **With cooperation**. In this context the legitimate user is asked to place his finger on a moldable and stable material in order to obtain the negative of the fingerprint. In a posterior step the gummy finger is recovered from the negative mold. The whole generation process is depicted in Fig. 2 and thoroughly described in [25].

– **Without cooperation**. In this case we recover a latent fingerprint that the user has unnoticedly left behind (on a CD in our experiments). The latent fingerprint is lifted using a specialized fingerprint development toolkit and then digitalized with a scanner. The scanned image is then enhanced through image processing and finally printed on a PCB from which the gummy finger is generated. The main steps of this non-cooperative process, first introduced in [8], are depicted in Fig. 3.

**Fig. 2** Process followed to generate fake fingerprints with the cooperation of the user: select the amount of moldable material (**a**), spread it on a piece of paper (**b**), place the finger on it and press (**c**), negative of the fingerprint (**d**). Mix the silicone and the catalyst (**e**), pour it on the negative (**f**), wait for it to harden and lift it (**g**), fake fingerprint (**h**)

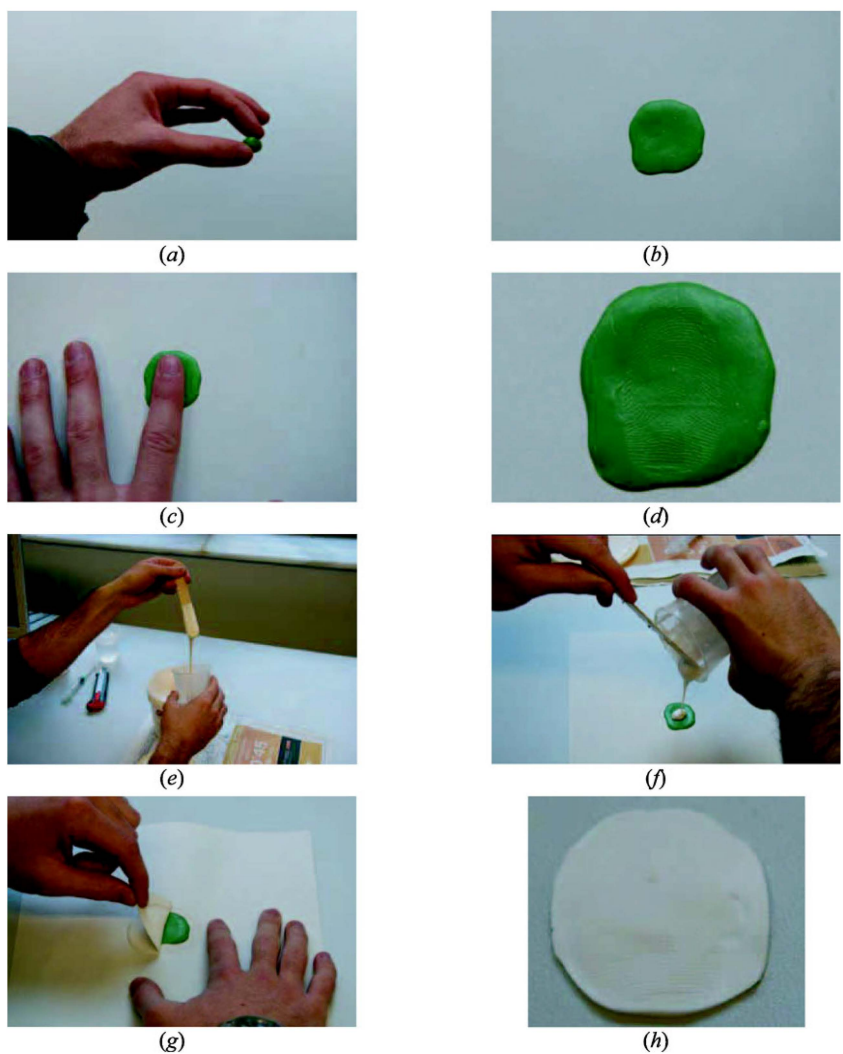

## 5 Database and experimental protocol

Experiments are carried out on a database comprising the index and middle fingers of both hands of 17 users ($17 \times 4 = 68$ different fingers). For each real finger, two fake imitations were generated following each of the procedures explained before (i.e., with and without the user's cooperation). Four samples of each fingerprint (fake and real) were captured in one acquisition session with: i) flat optical sensor Biometrika Fx2000 (512 dpi), ii) sweeping thermal sensor by Yubee with Atmel's Fingerchip (500 dpi), and iii) flat capacitive sensor by Precise Biometrics model Precise 100 SC (500 dpi). Thus, the database comprises 68 fingers $\times 4$ samples $\times 3$ sensors = 816 real image samples and as many fake images for each scenario (with and without cooperation). In order to ensure inter- and intra-class variability, samples of the same finger were not captured consecutively, following the methodology for biometric database acquisition developed in the project BioSec [26]. As described in Sect. 6, the images were labeled with the NFIS2 software package according to their quality level. Some good (top rectangle) and bad (bottom rectangle) quality samples of the database are depicted in Fig. 4.

Two different attack scenarios are considered in the experiments and compared to the normal operation mode of the system:

– **Normal Operation Mode (NOM)**: both the enrollment and the test are carried out with real fingerprints. This is used as the reference scenario. In this context the FAR (False Acceptance Rate) of the system is defined as the number of times an impostor using his own finger gains access to the system as a genuine user, which can be understood as the robustness of the system against a zero-effort attack. The same way, the FRR (False Rejection

**Fig. 3** Process followed to generate fake fingerprints without the cooperation of the user: latent fingerprint left on a CD (**a**), lift the latent fingerprint (**b**), scan the lifted fingerprint (**c**), enhance the scanned image (**d**), print fingerprint on PCB (**e**), pour the silicone and catalyst mixture on the PCB (**f**), wait for it to harden and lift it (**g**), fake fingerprint image acquired with the resulting gummy finger on an optical sensor (**h**)

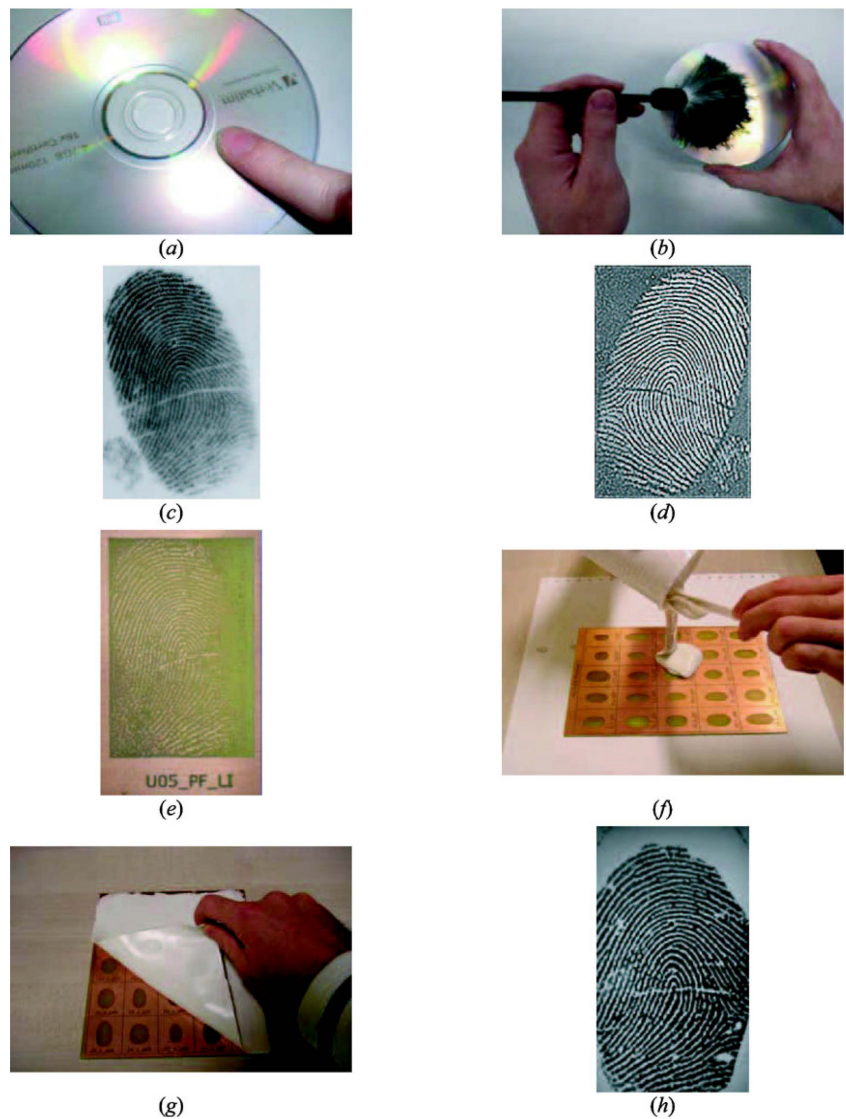

Rate) denotes the number of times a genuine user is rejected by the system.

– **Attack 1**: both the enrollment and the test are carried out with fake fingerprints. In this case the attacker enrolls to the system with the fake fingerprint of the genuine user and then tries to access the application with that same fake fingerprint. In this scenario an attack is unsuccessful (i.e., the system repels the attack) when an impostor enrolls to the system using the gummy fingerprint of a genuine user, and subsequently he is not able to access the system using that same fake fingerprint. Thus, the Success Rate of the attack in this scenario can be computed as: SR = 1 − FNMR, where FNMR is the False Non-Match Rate.

In order to compute the performance of the system in the normal operation mode, the following sets of scores are generated: i) for genuine tests all the 4 real samples of each user are matched against each other avoiding symmetric match-ings ($(4 \times 3)/2 = 6$ scores per user), which leads to $6 \times 68 = 408$ genuine scores, and ii) for impostor tests each of the four samples of every user are matched with all the samples of the remaining users in the database avoiding symmetric matchings, resulting in $(67 \times 4 \times 4 \times 68)/2 = 36,448$ impostor scores.

Similarly, in order to compute the FNMR in attack 1, all the 4 fake samples of each user are compared with each other avoiding symmetric matchings, resulting in a total 408 scores for each scenario (cooperative and non-cooperative).

– **Attack 2**: the enrollment is achieved using real fingerprints, and tests are carried out with fake fingerprints. In this case the genuine user enrolls with her fingerprint and the attacker tries to access the application with the corresponding gummy fingerprint. A successful attack is accomplished when the system confuses a fake fin-gerprint with its corresponding genuine fingerprint, i.e., SR = FMR where the FMR is the False Match Rate.

**Fig. 4** Examples of good (**A**) and bad (**B**) quality images of the database used in the direct attacks evaluation (available at http://atvs.ii.uam.es). In both charts real images acquired with the optical, capacitive, and thermal sensor, are shown in the *top row*. Their respective fake images generated with cooperation are shown in the *middle row*, and without cooperation in the *bottom row*

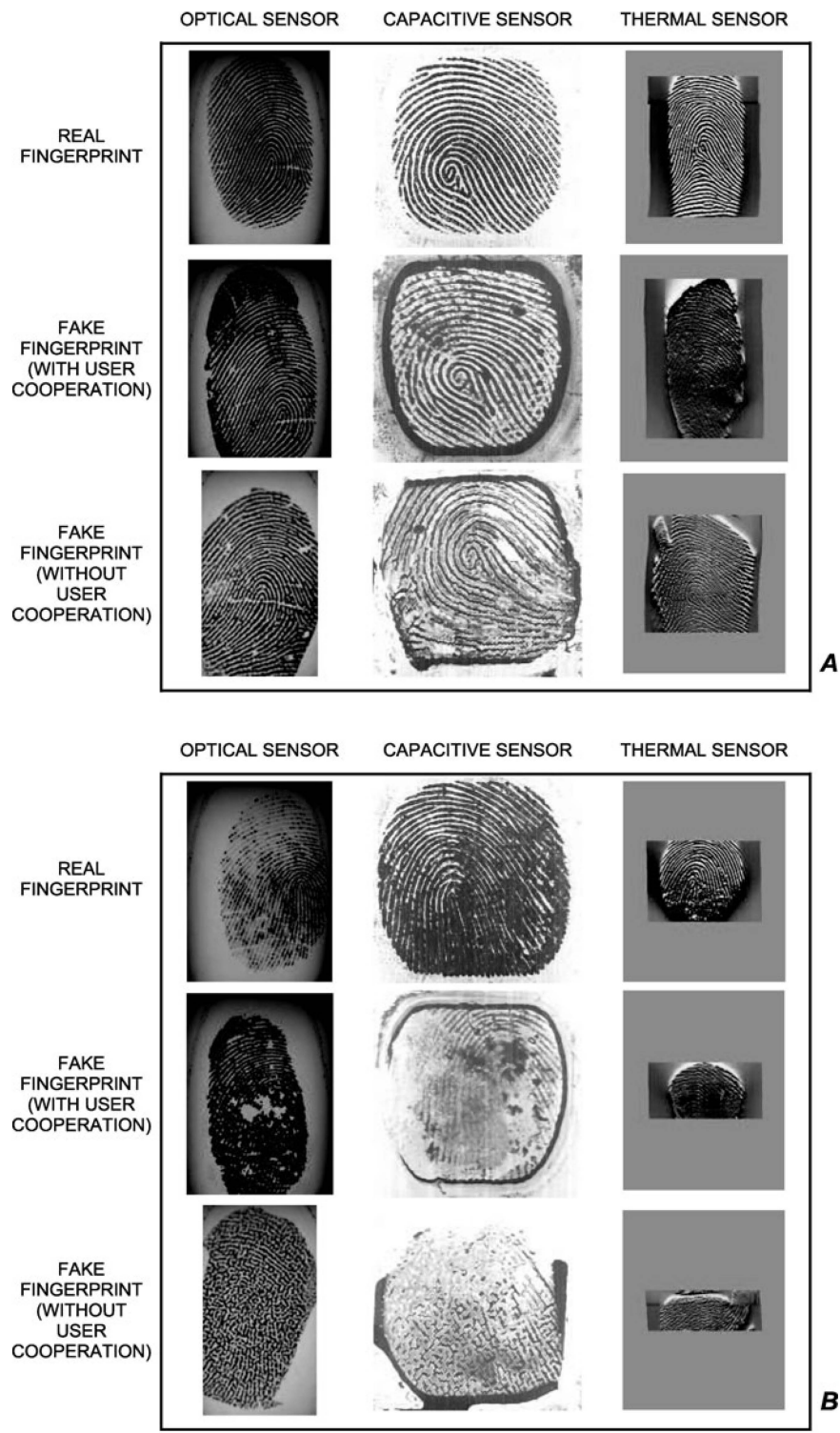

In this last scenario, only the impostor scores are computed matching all 4 original samples of each user with all 4 fake samples which results in 16 × 68 = 1,088 impostor scores for each scenario considered.

This experimental protocol was followed independently for the three sensors and the two kinds of fake fingerprints (cooperative and non-cooperative users).

## 6 Results

In Fig. 5 the DET curves of the two evaluated systems are shown for all the three sensors used in the experiments. The top two rows correspond to attacks carried out with fake fingerprints generated with the cooperation of the user, and the bottom rows without his cooperation. These results will be furthered analyzed in Sect. 6.2 and Sect. 6.3.

In addition to the evaluation of the attacks performance, we analyze the quality of the database samples comparing real and fake images. Then we study the success rates of the two attack scenarios compared to the performance at the normal operation mode.

### 6.1 Quality measures

The NFIS2 software package provides an application which estimates the quality of fingerprint images [27, 28]. This tool labels the samples from 1 to 5, being 1 the highest quality and 5 the lowest. In Fig. 6 we show the quality distribution of the image databases used in the experiments, both real and fake (with and without cooperation), for the three sensors used in the acquisition. We can see that the quality of the real samples (white bars) in all sensors is very high, being most of the samples concentrated in the first two levels. The quality of the fake images acquired with the optical sensor is acceptable (groups 2 and 3), while the fake images of the thermal and capacitive sensors are of very low quality (groups 3, 4, and 5).

These differences in the fake fingerprints quality is due to the three technologies used. The optical sensor is based on refraction effects of the light which take place in a similar way both in the skin and in the silicone of the gummy fingers, which leads to good fake images. On the other hand, the thermal sensor measures the difference in temperature between ridges and valleys which is nonexistent in the silicone, so, although the gummy fingers were heated up before being placed on the sensor, the resulting images are of poor quality. Similarly, the capacitive sensor is based on electrical measures, thus the silicone fingers had to be damped with a conductive substance in order to acquire the samples, which lead to low quality images.

Although the non-cooperative process to generate gummy fingers takes more steps (where the original fingerprint information might be degraded) than the cooperative procedure, the quality of the final fake images between both fake generation procedures only decreases significantly when acquired with the capacitive sensor. With the optical sensor the quality is just slightly worse for the non-cooperative process, while in the thermal sensor non-cooperative samples present even a better quality than those generated with the cooperation of the user. As will be shown in the next sections, these quality measures have a strong influence in the performance of the attacks evaluated.

### 6.2 NIST system evaluation

In Table 1(a) we show the Success Rate (SR) of the direct attacks against the NIST minutiae-based system at three different operating points. The decision threshold is fixed to reach a FAR = $\{0.1, 1, 10\}$ in the normal operation mode (NOM), and then the success rate of the two proposed attacks in analyzed in the two attack scenarios (with and without cooperation) for the three acquisition sensors.

The statistical significance of the results presented in Table 1 has been assessed using the model-based approach proposed in [29] using the Beta-Binomial distribution family. This method, specifically designed for the statistical analysis of biometric related problems, models the correlation between the multiple biometric acquisitions as well as accounts for varying FRR and FAR values for different subjects. The algorithm gives a non-symmetric confidence interval for the mean error rate at a given threshold value. In Table 1 the success rate of the two proposed attacks is given (in bold) together with (below in brackets) its 95% confidence interval (i.e., the probability of the success rate being within that interval is 0.95). Other works discussing the problem of statistical assessment in specific biometric studies can be found in [30–32].

**Attacks with cooperation**

When the optical sensor is used, due to the good quality samples acquired, the SR increases to reach over 65% in all of the operating points considered for attack 2 (the intruder tries to access the system with the gummy finger of a correctly enrolled user). On the other hand, the fake images captured with the thermal sensor show very little discriminant capacity, which leads to a very similar performance of the system against random impostors (FAR in NOM) and the SR of attack 2 for all the operating points studied. When the capacitive sensor is used, the system shows more resistance to the attacks than with the optical sensor, but it is more vulnerable than when the thermal technology is deployed (as corresponds to the intermediate quality level of the fake samples captured). The same effect can be observed in attack 1: as the quality of the fake samples increases (from the thermal to the optical sensor) the system becomes more vulnerable to the attacks.

**Attacks without cooperation**

In this scenario the samples captured with the thermal sensor present a higher quality than those acquired with the capacitive sensor and thus the SR of the attacks is lower in the latter case. We can also see that the performance of the attacks carried out using the optical sensor is lower when considering non-cooperative samples (compared to the samples generated with the cooperation of the user), as corresponds to a lower quality of the images.

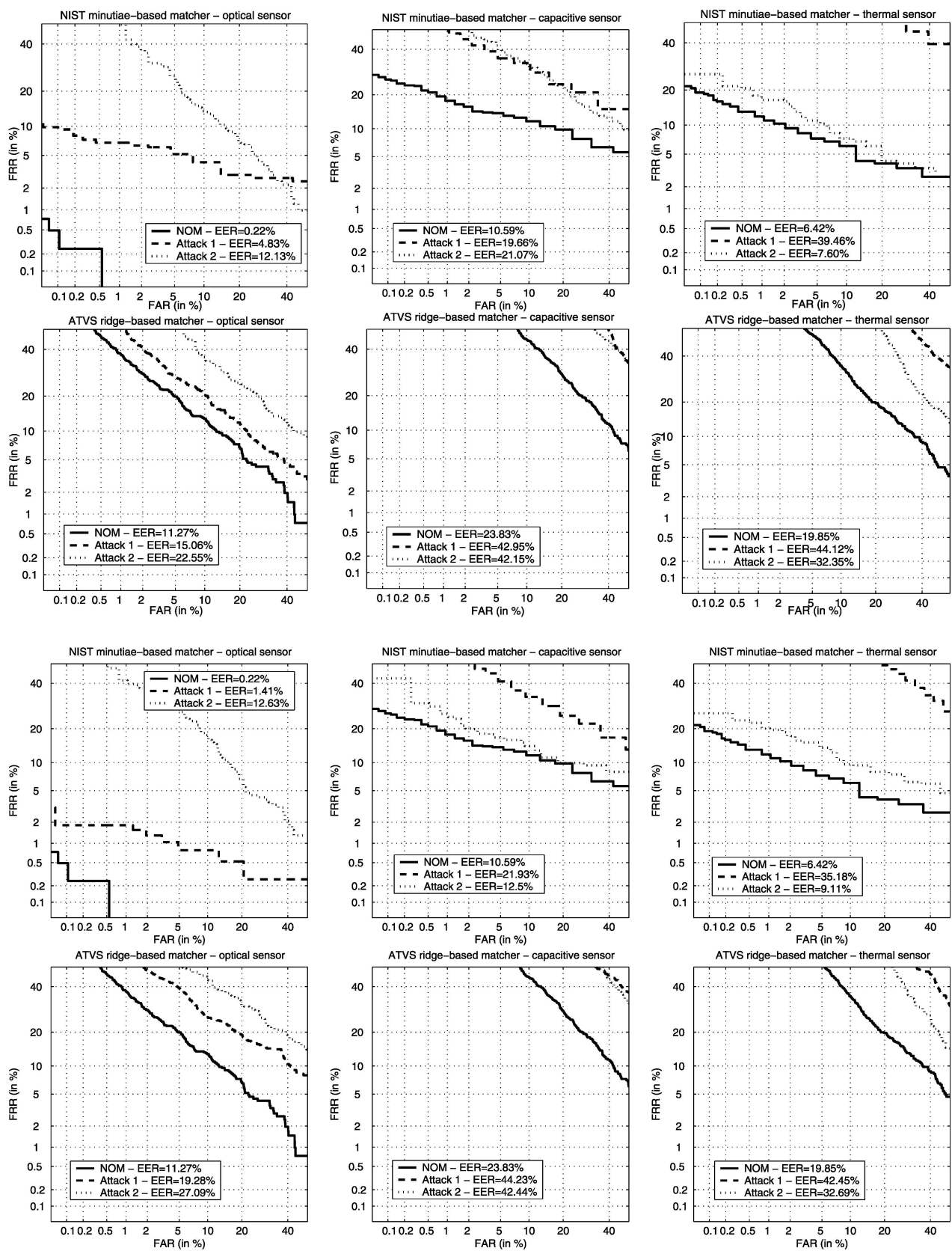

**Fig. 5** DET curves of the minutiae- and ridge-based systems for the three sensors used in the experiments (left to right: optical, capacitive and thermal). The *top two rows* correspond to attacks with cooperative users and the *bottom rows* with non-cooperative users

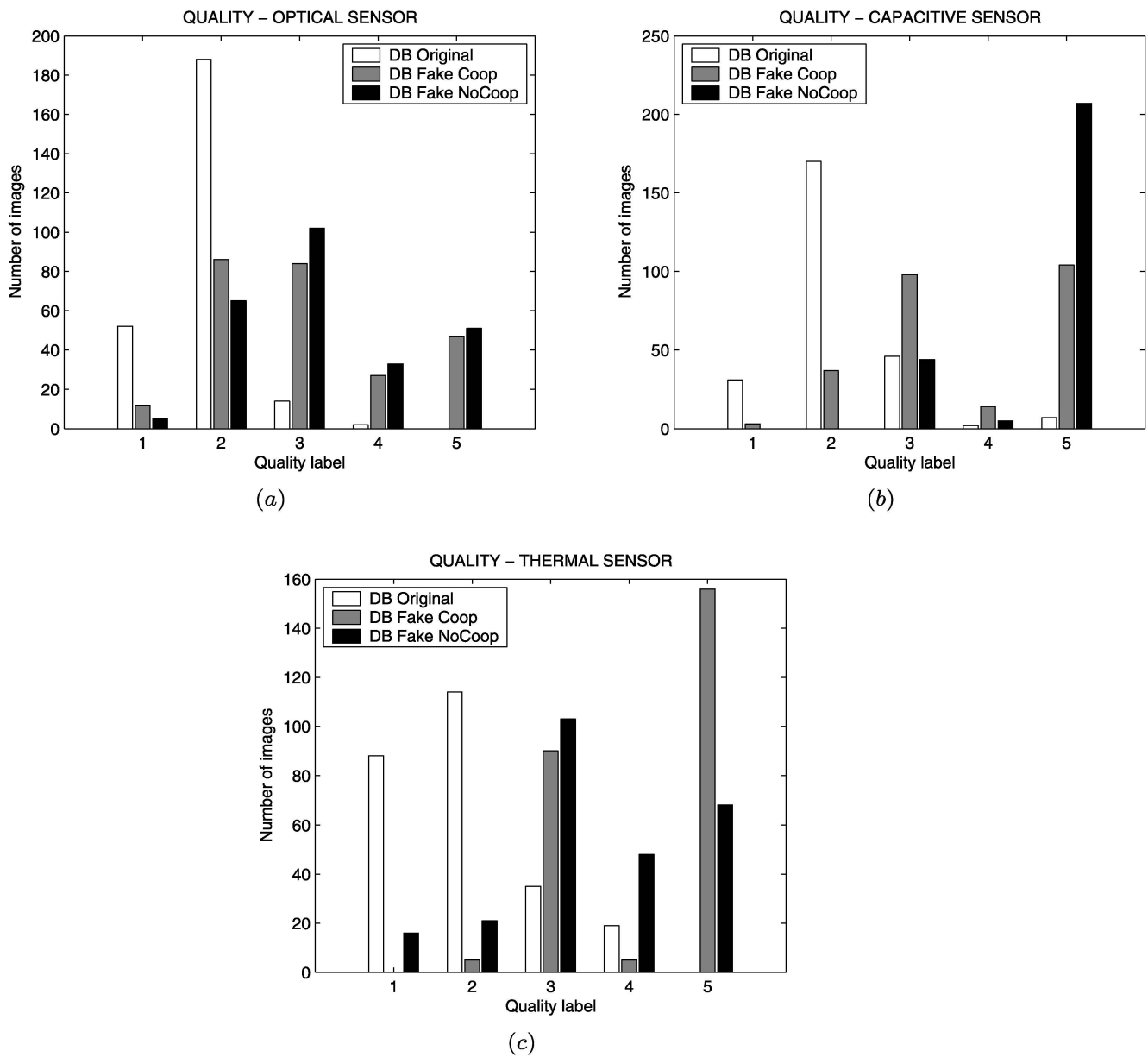

**Fig. 6** Quality distribution of the image databases (*1* corresponds to the highest quality, and *5* to the lowest), captured with the optical sensor (**a**), capacitive sensor (**b**), and thermal sweeping sensor (**c**)

On this basis, we can conclude that there exists a clear correlation between the quality of the fake fingerprint samples and the robustness against direct attacks of the NIST verification system: the better the image quality of the captured fake fingerprints, the higher the success rate of the attacks.

### 6.3 Ridge based system evaluation

In Table 1(b) we show the Success Rates of the attacks (SR) for the ridge-based system in an analog way to those presented in Table 1(a) for the NIST system.

**Attacks with cooperation**

In this case the difference between the robustness against random impostors (FAR in NOM) and the SR of attack 2 when using the optical sensor is significantly smaller than in the minutiae-based system. In addition, there are no noticeable differences in the success rate of the attacks between the three sensors used.

**Attacks without cooperation**

We observe that the SR is specially high for attack 1 with the thermal sensor, while specially low for attack 2 with the

**Table 1** Evaluation of the NIST and ridge-based systems to direct attacks with (Coop) and without (NoCoop) the cooperation of the user. NOM refers to the system normal operation mode and SR to the success rate of the attack (in bold). Attack 1 and 2 correspond to the attacks defined in Sect. 5 (enrollment/test with fakes/fakes for attack 1, and genuine/fakes for attack 2). The 95% confidence interval of the SR appears in brackets

(a) Performance of the attacks on the NIST minutiae-based system

|  | NOM | | Attack 1 | | Attack 2 | |
|---|---|---|---|---|---|---|
|  | FAR (%) | FRR (%) | SR (%) | | SR (%) | |
|  | | | Coop | NoCoop | Coop | NoCoop |
| Optical | 0.1 | 0.25 | **98.65** | **98.17** | **89.73** | **83.41** |
|  | | | (97.03,99.66) | (96.70,99.65) | (87.54,91.36) | (81.22,84.93) |
|  | 1 | 0 | **99.01** | **98.69** | **93.15** | **86.52** |
|  | | | (97.68,1) | (97.37,1) | (91.30,95.44) | (84.36,89.11) |
|  | 10 | 0 | **99.71** | **99.47** | **98.20** | **93.48** |
|  | | | (98.85,1) | (98.76,1) | (97.04,99.73) | (91.92,95.34) |
| Capacitive | 0.1 | 25 | **34.58** | **27.86** | **18.34** | **2.12** |
|  | | | (30.01,38.92) | (22.63,30.12) | (15.92,21.33) | (0.85,3.15) |
|  | 1 | 16 | **49.32** | **44.27** | **25.63** | **2.98** |
|  | | | (46.51,53.02) | (40.84,47.62) | (22.31,29.03) | (4.51,1.24) |
|  | 10 | 11 | **75.30** | **70.83** | **42.74** | **11.08** |
|  | | | (72.56,78.23) | (66.16,73.26) | (39.45,46.87) | (8.02,14.57) |
| Thermal | 0.1 | 18 | **8.09** | **43.75** | **0.81** | **0.78** |
|  | | | (5.88,10.92) | (39.65,46.17) | (0,1.69) | (0,1.65) |
|  | 1 | 11 | **13.46** | **73.17** | **4.95** | **4.68** |
|  | | | (9.25,17.62) | (69.84,77.65) | (2.53,7.04) | (2.45,6.92) |
|  | 10 | 6 | **46.95** | **89.06** | **18.32** | **21.61** |
|  | | | (40.99,49.35) | (85.10,93.14) | (16.05,21.69) | (17.51,25.78) |

(b) Performance of the attacks on the ridge-based system

|  | NOM | | Attack 1 | | Attack 2 | |
|---|---|---|---|---|---|---|
|  | FAR (%) | FRR (%) | SR (%) | | SR (%) | |
|  | | | Coop | NoCoop | Coop | NoCoop |
| Optical | 0.1 | 61 | **56.31** | **23.43** | **1.23** | **0.78** |
|  | | | (52.62,59.87) | (19.28,27.59) | (0.02,2.64) | (0,1.25) |
|  | 1 | 36 | **78.65** | **44.01** | **11.23** | **5.73** |
|  | | | (72.69,83.11) | (39.22,48.79) | (7.95,13.98) | (3.19,8.27) |
|  | 10 | 13 | **92.71** | **76.04** | **44.84** | **37.50** |
|  | | | (89.87,94.32) | (71.69,80.39) | (40.36,46.62) | (32.57,42.43) |
| Capacitive | 0.1 | 96 | **42.96** | **54.12** | **8.13** | **10.41** |
|  | | | (39.93,46.34) | (50.02,57.51) | (6.07,10.79) | (7.64,14.47) |
|  | 1 | 78 | **76.82** | **82.29** | **16.21** | **19.53** |
|  | | | (71.59,79.35) | (78.74,85.84) | (14.09,20.07) | (15.38,23.67) |
|  | 10 | 35 | **90.46** | **98.43** | **43.78** | **59.37** |
|  | | | (87.53,92.88) | (97.24,99.63) | (40.32,48.41) | (53.26,65.48) |
| Thermal | 0.1 | 95 | **46.34** | **52.34** | **2.16** | **0.78** |
|  | | | (41.24,48.93) | (48.43,56.25) | (1.01,3.65) | (0,1.65) |
|  | 1 | 82 | **85.31** | **90.62** | **11.54** | **7.29** |
|  | | | (82.86,88.43) | (87.82,93.42) | (8.14,13.97) | (4.68,9.89) |
|  | 10 | 45 | **93.47** | **99.15** | **32.82** | **25.02** |
|  | | | (92.16,95.17) | (98.32,1) | (29.61,36.87) | (20.70,29.29) |

optical sensor. Also as has been observed in the attacks with cooperation, the difference in the attacks performance between the three sensors is much lower than in the NIST system.

Thus, we can conclude that the ridge-based system is more robust to variations in the fingerprints quality, and less vulnerable to direct attacks with good quality fake images than the minutiae-based system from NIST.

## 7 Conclusions

A systematic and replicable evaluation of the vulnerabilities of fingerprint-based recognition systems to direct attacks has been presented. The attacks were performed on the NIST minutiae-based system and a proprietary ridge feature-based system, and were studied on a database of real and fake samples from 68 fingers, generated with and without the cooperation of the legitimate user, captured with three different sensors (optical, thermal and capacitive), with 4 impressions per finger. The resulting fingerprint image databases will be publicly available at http://atvs.ii.uam.es.

Two different attacks were considered, namely: i) enrollment and test with gummy fingers, and ii) enrollment with a real fingerprint and test with its corresponding fake imitation. Statistically significant results on the performance of the attacks were reported and compared to the normal operation mode of the system.

The results show that, when considering the minutiae-based system, the attacks success rate is highly dependent on the quality of the fake fingerprint samples: the better the image quality of the captured fake fingerprints, the lower the robustness of the system against the two studied attacks. The ridge-based system proved to be more robust to high quality fake images and, in general, to variations in fingerprint image quality.

Liveness detection procedures [15], or multimodal authentication architectures [33], are possible countermeasures against this type of fraudulent actions.

**Acknowledgements** This work has been supported by Spanish Ministry of Defense, and the TEC2006-13141-C03-03 project of the Spanish MCyT. J. G. is supported by a FPU Fellowship from the Spanish MEC. J. F. is supported by a Marie Curie Fellowship from the European Commission. F. A.-F. is supported by a Juan de la Cierva Fellowship from the Spanish MICINN. The authors would also like to thank Daniel Hernandez-Lopez and Guillermo Gonzalez de Rivera for their valuable development work.

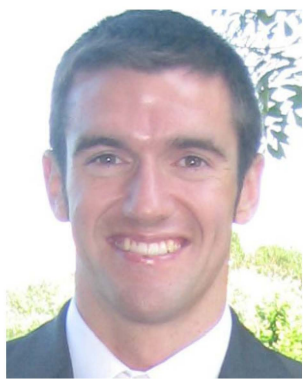

**J. Galbally** received the MSc in electrical engineering in 2005 from the Universidad de Cantabria, and the PhD degree in electrical engineering in 2009, from Universidad Autonoma de Madrid, Spain. Since 2006 he is with Universidad Autonoma de Madrid, where he is currently working as an assistant researcher. He has carried out different research internships in worldwide leading groups in biometric recognition such as BioLab from Universita di Bologna Italy, IDIAP Research Institute in Switzerland, or the Scribens Laboratory at the École Polytecnique de Montréal in Canada. His research interests are mainly focused on the security evaluation of biometric systems, but also include pattern and biometric recognition, and synthetic generation of biometric traits. He is actively involved in European projects focused on vulnerability assessment of biometrics (e.g., STREP Tabula Rasa) and is the recipient of a number of distinctions, including: IBM Best Student Paper Award at ICPR 2008, and finalist of the EBF European Biometric Research Award.

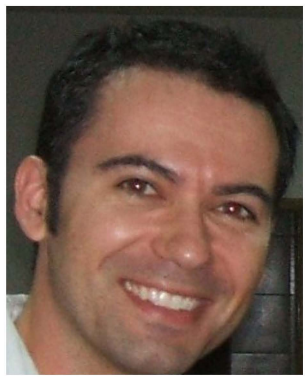

**J. Fierrez** received the M.Sc. and the Ph.D. degrees in electrical engineering in 2001 and 2006, respectively, both from Universidad Politecnica de Madrid, Spain. Since 2004 he is with Universidad Autonoma de Madrid, where he is currently working as an assistant researcher. Starting in early 2007, he will be a visiting researcher at Michigan State University for 2 years, under a Marie Curie Fellowship. His research interests include signal and image processing, pattern recognition and biometrics. He is actively involved in European projects focused on biometrics (e.g. Biosecure NoE) and is the recipient of a number of distinctions, including: best poster award at AVBPA 2003, 2nd best signature verification system at SVC 2004, Rosina Ribalta award to the best Spanish PhD proposal in 2005, Motorola best student paper at ICB 2006, and EBF European Biometric Industry Award 2006.

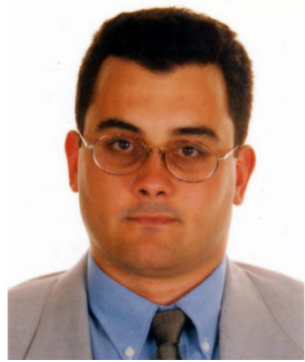

**F. Alonso-Fernandez** received the M.S. degree in 2003 with Distinction and the Ph.D. degree "cum laude" in 2008, both in Electrical Engineering, from Universidad Politecnica de Madrid (UPM), Spain. He is currently the recipient of a Juan de la Cierva postdoctoral fellowship of the Spanish Ministry of Innovation and Science. His research interests include signal and image processing, pattern recognition and biometrics. He has published several journal and conference papers and he is actively involved in European projects focused on biometrics (e.g., Biosecure NoE, COST 2101). He has participated in the development of several systems for a number of biometric evaluations (e.g. SigComp 2009, LivDet 2009, BMEC 2007). Dr. Alonso-Fernandez has been invited researcher in several laboratories across Europe.

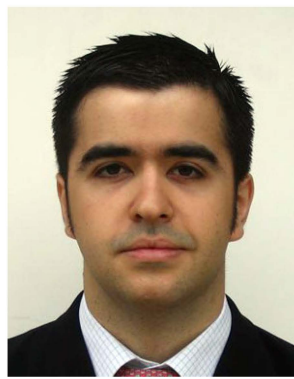

**M. Martinez-Diaz** received the MSc degree in Telecommunication Engineering in 2006 from Universidad Autonoma de Madrid, Spain. He worked as an IT strategy consultant until 2007. Currently he is working in the Technology area of Vodafone Spain. Since 2005 he is with the Biometric Recognition Group - ATVS at the Universidad Autonoma de Madrid, where he is collaborating as student researcher pursuing the PhD degree. His research interests include biometrics, pattern recognition and signal processing primarily focused on signature verification.